\documentclass[10pt]{article}
\usepackage[preprint]{rlj}
\usepackage{amssymb}
\usepackage{mathtools}
\usepackage{mathrsfs}
\usepackage{graphicx}
\usepackage{subcaption}
\usepackage[space]{grffile}
\usepackage{wrapfig}
\usepackage{url}
\usepackage{lipsum}
\usepackage{hanging}
\newcommand{\hanging}[1]{\hangpara{2em}{1}#1}

\title{Physical Atari: A Robust and Accessible Platform for Real-time Reinforcement Learning on Robots}

\setrunningtitle{Physical Atari}

\author{Khurram Javed\textsuperscript{1}, Joseph Modayil\textsuperscript{1,3}, Gloria Kennickell\textsuperscript{1}, \\ Richard S. Sutton\textsuperscript{1,2}, John Carmack\textsuperscript{1}}

\emails{\{kjaved,joseph,gloria,rssutton,johnc\}@keenagi.com}

\affiliations{
$^{1}$\textbf{Keen Technologies}
\\
$^{2}$\textbf{University of Alberta, Canada}
\\
$^{3}$\textbf{Openmind Research Institute}
}

\contribution{

    We introduce \textit{Physical Atari}, a low-cost, reliable platform for studying real-time reinforcement learning on physical robots. The platform consists of \textit{the Robotroller}, a robot that actuates a standard Atari controller, and \textit{the Atari Devbox}, a system that renders the game frame and the reward signal at 60 FPS from the Arcade Learning Environment.
    }
    {
    The Arcade Learning Environment (ALE) has been instrumental in advancing reinforcement learning in simulation. However, real-world robotics introduces challenges such as latency, wear, and non-stationarity that are ignored in simulation. Physical Atari brings the richness of the ALE benchmark to the physical world and can be used to study these challenges in a systematic way.
    }

\contribution{
    We demonstrate that policies trained on one robot perform significantly worse when transferred to an identically manufactured robot, highlighting the sensitivity of learned policies to minute physical variations. We further show that this performance gap can be bridged through continual on-robot learning.
    }
    {
    A common paradigm in robotics is to train policies in simulation or on a specific robot and deploy them broadly. This approach assumes that hardware variations are negligible or can be handled by robust policies. Our findings challenge this assumption, suggesting that even subtle distribution shifts between ``identical'' hardware can severely limit performance, underscoring the necessity of on-device adaptation. The goal of creating the Physical Atari platform was exactly to explore these types of questions.
    }

\summary{We built a robot called \textit{the Robotroller} that actuates an Atari CX40+ controller and a device called \textit{the Atari Devbox} that renders the game frame and the reward signal from the Arcade Learning Environment on a screen. The Robotroller and the Atari Devbox, together with an off-the-shelf camera and a desktop computer, constitute a system that can be used to study reinforcement learning algorithms in the physical world. We call the full system \textit{Physical Atari}. In this paper, we detail the key decisions that make Physical Atari a robust and accessible platform. To make the system robust, we designed the Robotroller so that all movement is done through bearings, which reduces wear. Additionally, we wrote software that monitors the state of the servos at a high frequency and intervenes to limit stress. To make the system accessible, we used affordable off-the-shelf components and parts that can be manufactured using consumer 3D printers. Physical Atari can be built for under \$1,000 and has been used for weeks of non-stop reinforcement learning experiments without any mechanical failures. We used it to validate that reinforcement learning algorithms can learn directly on robots and show that even small distribution shifts between learning and deployment can significantly degrade the performance of policies.
}

\begin{document}

\maketitle

\begin{abstract}
  We built a robot called \textit{the Robotroller} that actuates an Atari CX40+ controller and a device called \textit{the Atari Devbox} that renders the game frame and the reward signal from the Arcade Learning Environment on a screen. The Robotroller and the Atari Devbox, together with an off-the-shelf camera and a desktop computer, constitute a system that can be used to study reinforcement learning algorithms in the physical world. We call the full system \textit{Physical Atari}.
  
  In this paper, we detail the key decisions that make Physical Atari a robust and accessible platform. To make the system robust, we designed the Robotroller so that all movement is done through bearings, which reduces wear. Additionally, we wrote software that monitors the state of the servos at a high frequency and intervenes to limit stress. To make the system accessible, we used affordable off-the-shelf components and parts that can be manufactured using consumer 3D printers. Physical Atari can be built for under \$1,000 and has been used for weeks of non-stop reinforcement learning experiments without any mechanical failures. We used it to validate that reinforcement learning algorithms can learn directly on robots and show that even small distribution shifts between learning and deployment can significantly degrade the performance of policies. Our results underscore the importance of on-device adaptation for strong performance on robots. 
\end{abstract}

\section{Real-time Reinforcement Learning}

A reinforcement learning agent uses its sensors to observe aspects of its world, uses the sensory data and its internal state to make decisions, and uses its effectors to influence the environment with the aim of maximizing its reward rate. Reinforcement learning benchmarks implement the agent-environment interaction as a turn-based game (for example, see Brockman et al., 2016; Bellemare, Naddaf \& Bowling, 2013; Beattie et al., 2016; Mahmood et al., 2018), in which the environment waits for the agent's response before changing. This is different from problems in the physical world, in which the state of the environment continues to evolve as the agent makes a decision. We call the setting in which the environment evolves continuously \textit{real-time reinforcement learning}. Real-time reinforcement learning is not a novel setting. For example, Travnik et al. (2018) and Ramstedt \& Pal (2019) study this setting.

The real-time reinforcement learning setting poses unique challenges. An agent in this setting may want to make a decision quickly, to minimize the difference between its perception of the state of the environment and the actual state of the environment. It may want to use different amounts of compute for different decisions, making it possible to take deliberate decisions when the cost of a delay is low and reactive actions when a delay can be catastrophic.

\begin{figure}[t]
  \centering

  \includegraphics[width=\textwidth]{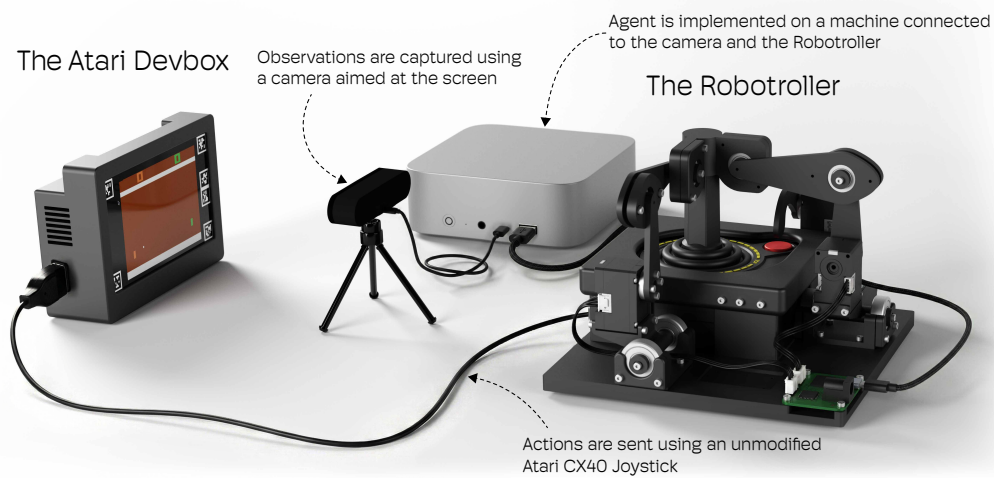}
  \caption{The overview of the Physical Atari platform. On the left is the Atari Devbox---a device that renders Atari 2600 games at 60 FPS, listens to the actions from the CX40+ controller, and displays AprilTags to communicate the reward signal and the location of the game screen. On the right is the Robotroller---a robot that actuates an unmodified CX40+ controller. A reinforcement learning agent that runs on a computer connected to the camera and the Robotroller captures the screen of the Atari Devbox using a camera and sends commands to the Robotroller to play the game.}
  \label{fig:overview}
\end{figure}

A direct application of the real-time reinforcement learning setting is robotics. In recent years, reinforcement learning has become a key paradigm for solving challenges in robotics. It has been applied to robotics in different ways. We describe three of them here.

One approach is to create a simulation of the robot and its environment and then use reinforcement learning algorithms to learn policies in the simulated environments (see Abbeel et al., 2006; OpenAI et al., 2019). These policies are then deployed to the robot. A second approach is to get human operators to control the robots and collect trajectories (see Mandlekar et al., 2018; Zhao et al., 2023). These trajectories are then used by an offline-RL algorithm to learn a policy that is deployed to the robot. A third, less common approach is to learn directly on the robots (see Benbrahim et al., 1992; Haarnoja et al., 2019; Smith et al., 2023; Wu et al., 2023). Two key advantages of learning directly on the robots are that there is no need for a simulator or human data and there is no distribution shift between the learning and deployment environments. The focus of this paper is the third approach.

There is no platform for studying reinforcement learning directly on robots---the third approach mentioned above---that is both reliable and accessible. We wanted to build one such platform. We built a platform called \textit{Physical Atari} that is \textit{accessible}, \textit{reliable}, \textit{easy to use}, and \textit{versatile}. Here accessible means that it is built using affordable and easily available components, reliable means that it can be used for long experiments without mechanical failures, easy to use means that it can be used for long periods without requiring interventions, and versatile means that it supports a wide range of tasks.

The two key components of Physical Atari are \textit{the Robotroller} and \textit{the Atari Devbox}. The Robotroller is a robot that actuates an unmodified commercially available Atari CX40+ controller. The Atari Devbox is a device with a 5-inch display that renders the game screen and the reward signals of games from the Arcade Learning Environment (ALE) (Bellemare et al., 2013).

\begin{wrapfigure}{R}{0.5\textwidth}
  \centering
  \includegraphics[width=0.5\textwidth]{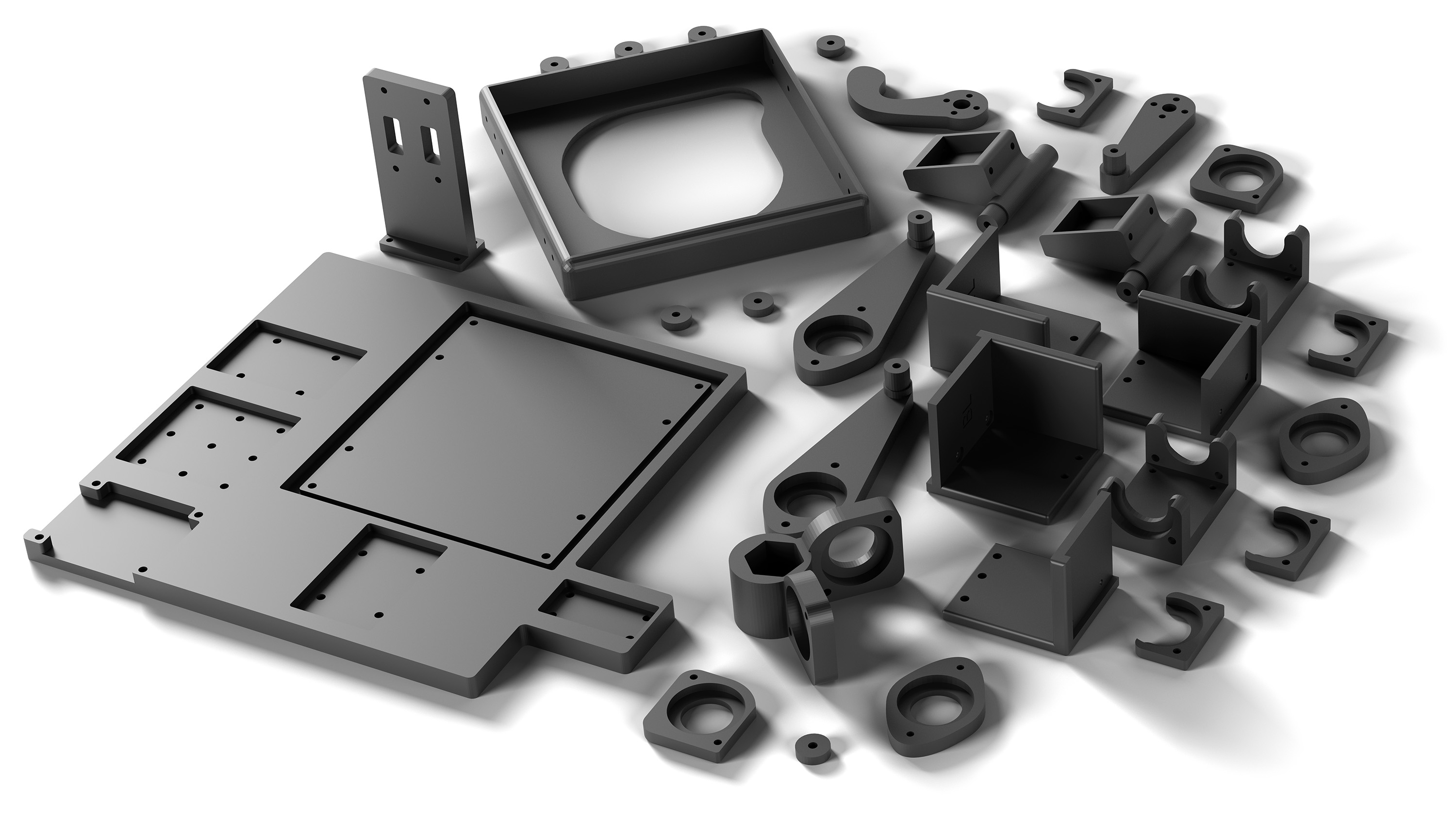}
  \caption{The components of the Robotroller that are 3D printed. On a consumer Bambu Lab P1S printer, all these parts can be printed in around 12 hours.}
  \label{fig:parts}
\end{wrapfigure}

We picked games from ALE as the tasks for our platform because the performance of reinforcement learning algorithms in simulation on these games is well understood, and these games have proven useful for reinforcement learning research. ALE has been used in several seminal works, such as DQN (Mnih et al., 2015), Rainbow DQN (Hessel et al., 2018), MuZero (Schrittwieser et al., 2020), Data-efficient Rainbow (Van Hasselt et al., 2019), and BBF (Schwarzer et al., 2023). Physical Atari brings the richness of the ALE benchmark to the physical world and can be used to study challenges, such as time delays and non-stationarities, that are ignored in simulation.

\section{The Physical Atari Platform}
\autoref{fig:overview} shows an overview of the Physical Atari platform. The platform consists of a camera that is pointed at the Atari Devbox, a computer that is connected to the camera, the Robotroller that is connected to the computer, and a CX40+ controller that is connected to the Atari Devbox.
\begin{wrapfigure}[18]{R}{0.5\textwidth}
  \centering
  \includegraphics[width=0.5\textwidth]{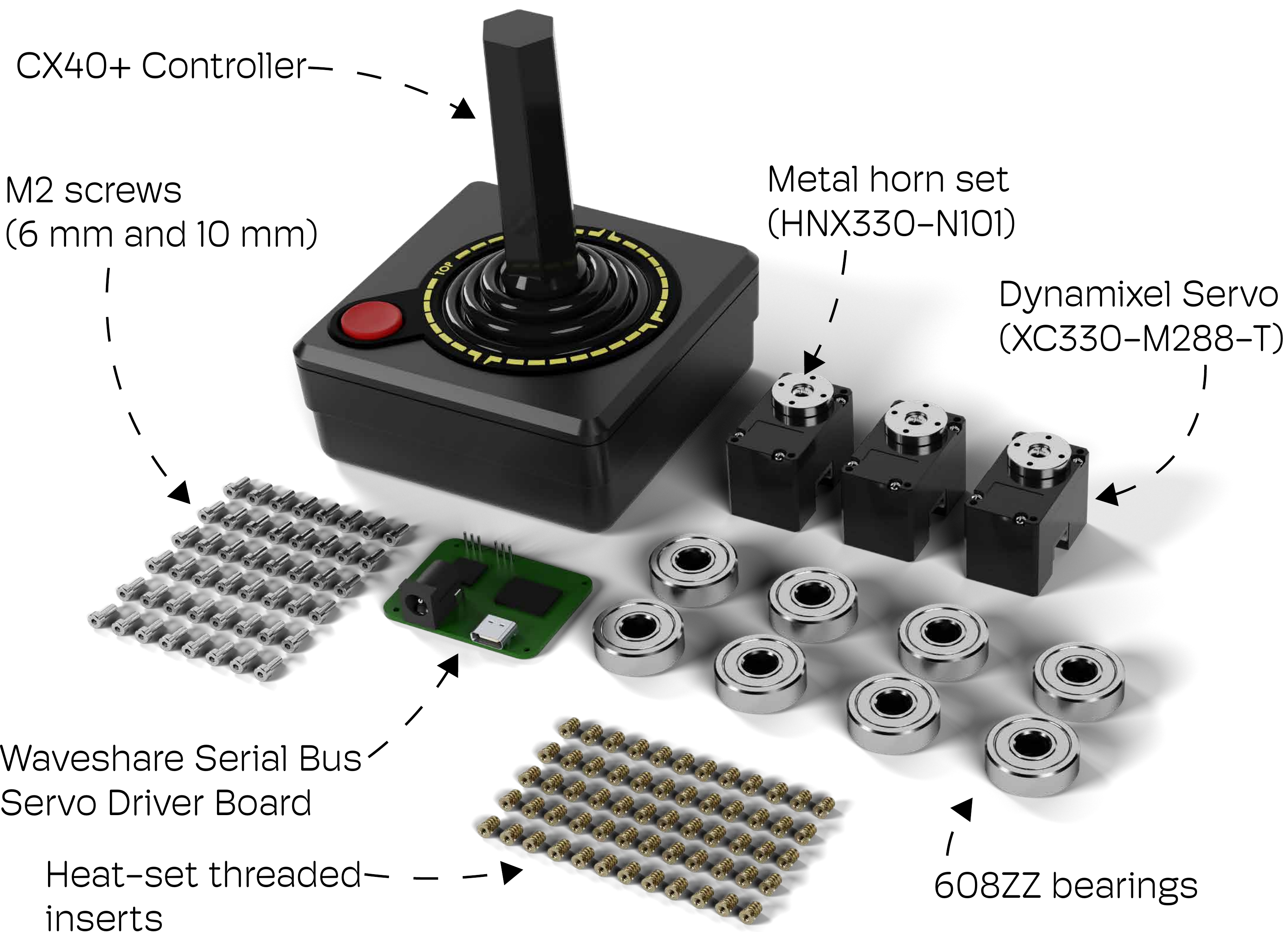}
  \caption{The components of the Robotroller that have to be purchased. These include screws, bearings, servos, electronics, and threaded inserts. The total cost of the parts is around \$400.}
  \label{fig:partstobought}
\end{wrapfigure}

The computer uses the camera as its sensor, and it sends one of the eighteen discrete actions used by ALE to the Robotroller. The actions are executed by moving the servo motors to preprogrammed positions.

The camera is the Razer Kiyo Pro, which is an off-the-shelf camera capable of streaming uncompressed video. Using an uncompressed data stream reduces latency. The camera is connected to the computer using a USB 3.1 connection. We used a Framework desktop with an AMD Ryzen AI Max+ 395 chip as our computer; other computers that support USB 3.1 should work fine.
\subsection{The Robotroller}
\begin{figure}[h]
  \centering
  \includegraphics[width=\textwidth]{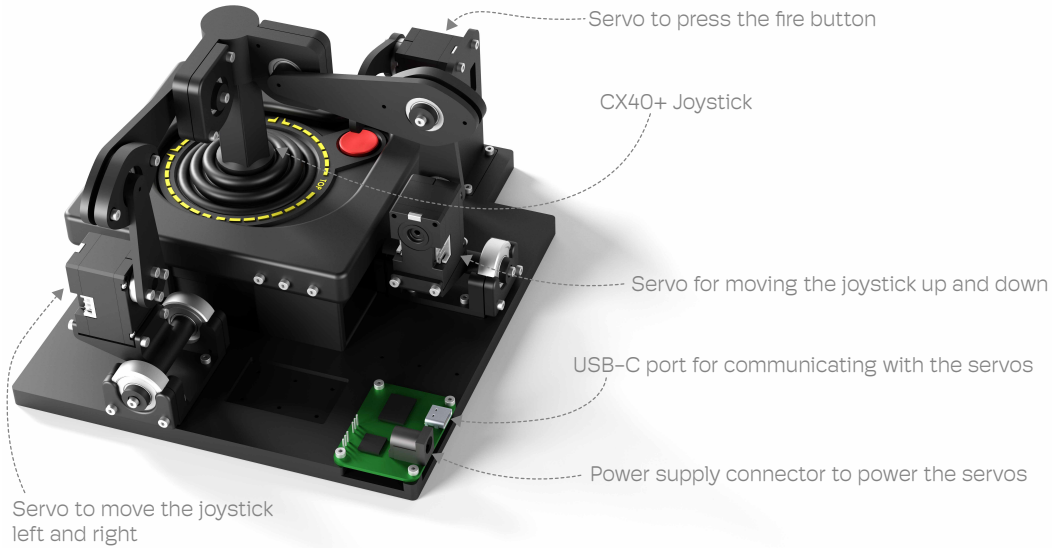}
  \vspace{-35pt}
  \caption{The Robotroller is a robot that reliably actuates an unmodified CX40+ controller using three servo motors. It uses the Dynamixel XC-330 servo motors and a Waveshare PCB to communicate with the servo motors over a serial connection via a USB-C port. To minimize wear, the Robotroller uses 608ZZ bearings in all moving parts.}
  \label{fig:robotroller}
\end{figure}

The Robotroller consists of two types of parts. The first type is custom parts designed in CAD software (Autodesk Fusion) that have to be manufactured. These are shown in \autoref{fig:parts}. We designed these parts so they can be printed on a consumer 3D printer, like the Bambu Lab P1S, using an inexpensive filament, such as PLA. The second type is parts that are commercially available. These include screws, bearings, servos, threaded inserts, and electronics. These are shown in \autoref{fig:partstobought}. The fully assembled Robotroller is shown in \autoref{fig:robotroller}.

\begin{figure}[h]
  \centering

  \includegraphics[width=\textwidth]{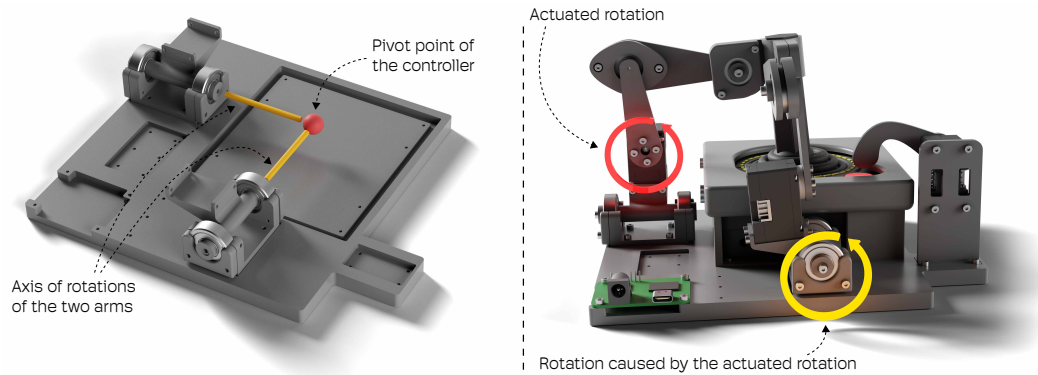}
  \vspace{-20pt}
  \caption{The key to making the Robotroller reliable is the design that uses bearings for all movements. This was a challenge to implement because as one arm of the robot pulls or pushes the joystick, the other arm has to move to accommodate the movement of the joystick. We deal with this by putting both arms on bearings whose axes of rotation are aligned with the pivot point of the joystick. The two axes meet exactly at the pivot point, as shown in the figure on the left. The figure on the right shows how rotation of one servo (axis of rotation shown as a red circle) causes the other arm to tilt (axis of rotation shown as a yellow circle).}
  \label{fig:minizeimpact}
\end{figure}

A key goal when designing the Robotroller was to ensure it can run for long periods without significant wear and tear. To achieve this goal, we used ball bearings for all moving parts of the system. We used 608ZZ bearings that are readily available at hardware stores. Using ball bearings reduces friction, which reduces wear that is typically associated with 3D-printed PLA parts. To transfer the torque of the servos to the joystick through ball bearings, we positioned the two servos that control the joystick so they can rotate about two axes that meet at the pivot point of the joystick. This configuration is illustrated in \autoref{fig:minizeimpact}.

When testing the Robotroller, we discovered that some screws would eventually come loose after extended use. We fixed this by using a thread-locking compound when assembling the Robotroller. We used \textit{Loctite 243 Threadlocker}.

After further testing, we discovered that the internal gears of the servos would wear out after extended use. This happened because in the initial design, we used the cheaper Dynamixel XL-330 servos, which use gears made of engineering plastic. We replaced the servos with the more expensive Dynamixel XC-330 servos that use metal gears. In our extended tests lasting several weeks of continuous use, none of our robots showed signs of wear in the metal gears.

The XC-330 servo uses metal gears, but its horn---the part used to connect the servo to external components---is made of engineering plastic. We found that in one build, this part broke. Dynamixel sells a metal replacement (HNX330-N101) for this part. We replaced the plastic horns with the metal horns on all three servos, which fixed the issue.

Once we made the Robotroller reliable, we noticed that the CX40+ controller wore out after extended use. This was surprising because the controller is used by people for long periods of time without issues. We traced the issue to the motion profile of the Robotroller, which was too aggressive and put unnecessary stress on the controller.

The servos use a PID controller for position-based control. The setpoints for positions are set manually so the servos can move to trigger any one of the eighteen actions of ALE when given a command. The PID controller has three key parameters---P gain, I gain, and D gain. By default, Dynamixel ships the servos with a large value of the P gain parameter, and zero values for the I gain and D gain parameters. We noticed that the large value of the P gain parameter created jerky movements, which likely put too much stress on the controller.

We made the motion of the servos smoother by changing the values of the PID controller parameters. We used a small value of the P gain parameter, a small value of the D gain parameter, and a large value of the I gain parameter. The exact values are in \autoref{app:robot_camera_hyperparameters}, and a video showing the movement of the Robotroller before and after tuning is \href{https://keenagi.com/research/physical-atari/pid_tuning.html}{here}.

The new values of the PID controller introduced another problem. If, for some reason, a servo did not reach its setpoint quickly, then the current would spike to a very large value and put the servo in a state that required a manual reboot. This was because of the large value of the I gain parameter, which integrates the error over time to generate a control signal. We fixed this by introducing a \textit{high-current reflex}---a preprogrammed behavior for the servos that is triggered if the current exceeds a certain threshold.
\begin{figure}[h]
  \centering
  \includegraphics[width=\linewidth]{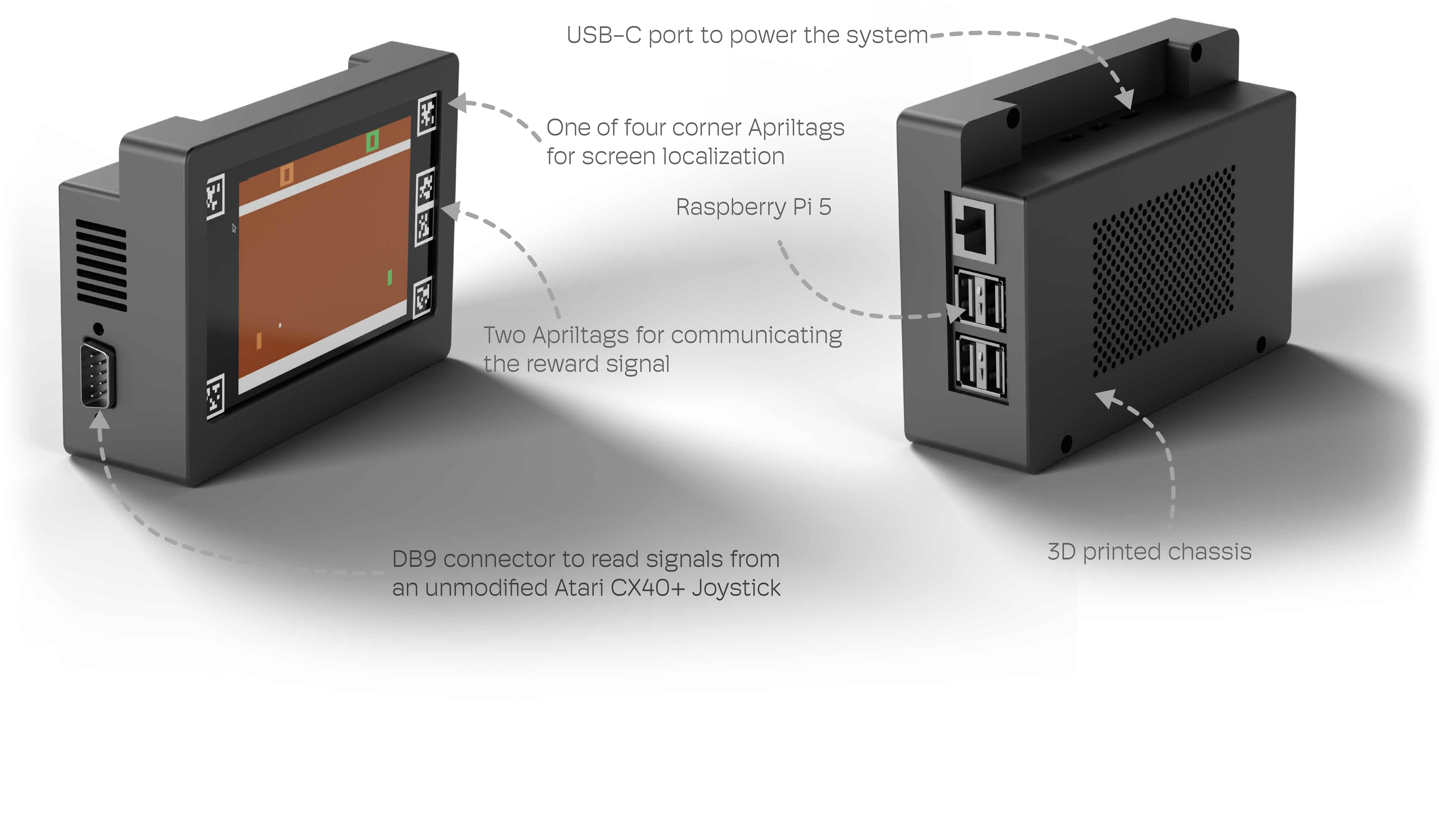}
  \vspace{-80pt}
  \caption{The Atari Devbox is a device that renders Atari 2600 games at 60 FPS, listens to the signals from the CX40+ controller, and displays AprilTags to communicate the reward signal and the location of the game screen. It uses the Arcade Learning Environment (ALE) running on a Raspberry Pi 5. To read signals from the CX40+ controller, it connects the controller to the GPIO pins of the Raspberry Pi 5. It uses an off-the-shelf Waveshare Display that connects to the Raspberry Pi 5 using its SPI interface. A C++ program running on the Raspberry Pi 5 reads the signals from the GPIO pins at a very high frequency and converts them to one of the eighteen discrete actions of the Arcade Learning Environment.}
  \label{fig:devbox}
\end{figure}

The high-current reflex is implemented in a high-frequency loop that continually checks the current draw of the servos. If at any point the current exceeds a threshold, the goal position of the servo is set to its current position, and its torque is disabled for one millisecond. This reflex acts as a safety mechanism to protect the servos from damaging themselves or entering a state that requires a manual reboot. We call this policy a reflex because there are numerous examples of reflexes in the human body that protect the body from damage in a similar way.\footnote{An example is the Golgi tendon reflex, a reflex to protect against too much tension on the tendons and muscles of the body.}

After updating the parameters of the PID controller and adding the high-current reflex, we used the Robotroller continuously for weeks with the same CX40+ controller without signs of significant wear or loss of function. We provide more details of the Robotroller in \autoref{app:cpp_library}, and the complete step-by-step instructions for building the Robotroller are \href{https://keenagi.com/research/physical-atari/robotroller.html}{here}.

\subsection{The Atari Devbox}
The Atari Devbox is a compact device that uses a Raspberry Pi 5 as its compute unit and a 5-inch Waveshare display. We designed a custom chassis in Autodesk Fusion that can be 3D-printed to house the components. Since the Raspberry Pi 5 lacks a built-in DB9 port required to connect the Atari CX40+ joystick, we connected a male DB9 connector directly to the GPIO pins of the Raspberry Pi 5. This connector is exposed through the chassis of the Atari Devbox. A picture of the Atari Devbox is shown in \autoref{fig:devbox}.

We wrote a program that reads the state of the GPIO pins with sub-millisecond latency and maps the state of the five wires of the CX40+ controller (Up, Down, Left, Right, Fire) to the eighteen discrete actions of the Arcade Learning Environment (ALE). The program then passes the action to the ALE environment and renders the resulting game frame to the screen at 60 FPS.

Beyond the game screen, the program renders a set of AprilTags~(Olson, 2011): four corner tags to help an agent localize the game screen from camera images, and a dynamic set of tags on one side of the screen to encode the reward signal.

The complete step-by-step instructions for building the Atari Devbox are available \href{https://keenagi.com/research/physical-atari/devbox.html}{here} and more details about the Atari Devbox are in \autoref{app:devbox}.

\section{Characterizing the Response Time of Physical Atari}
A physical system has a delay between receiving a stimulus and responding. Rendering a frame to the screen, capturing the screen with a camera, processing the image to get an action, and moving the controller to execute the action all take time.
\begin{figure}[t]
  \centering
  \includegraphics[width=\textwidth]{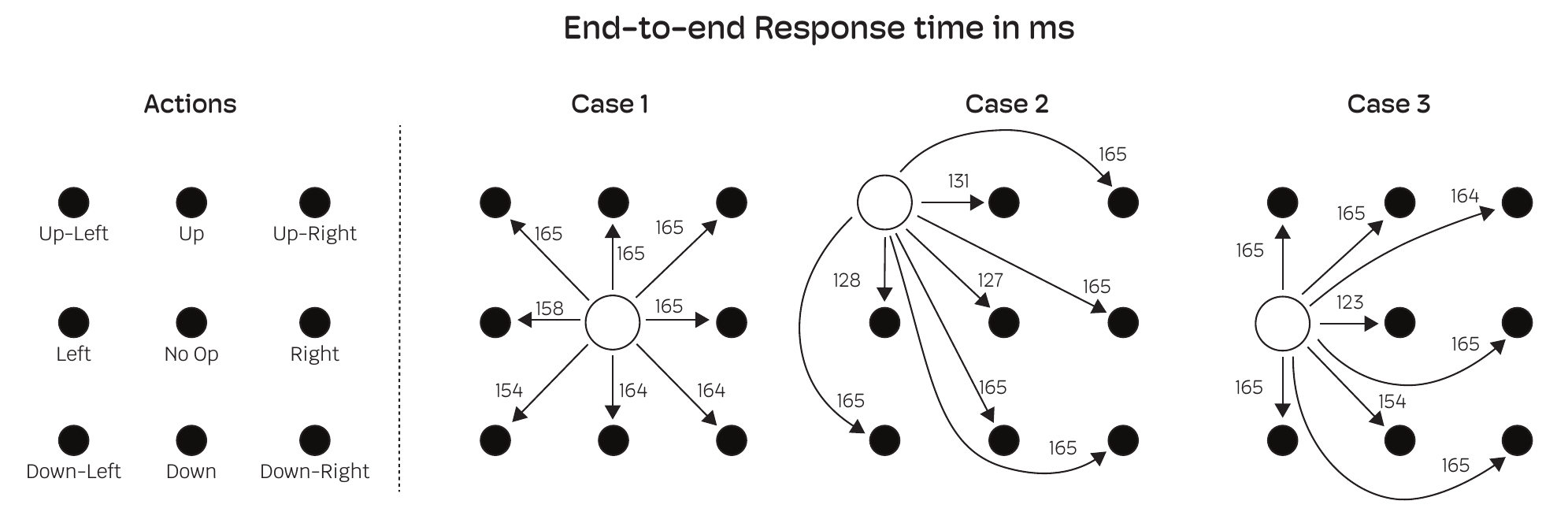}
  \caption{We measured the end-to-end response time of the Physical Atari platform. The response time is the total time required to render on the screen, capture the image with the camera, transfer the camera image to the computer, and move the joystick to the desired action. We plot the times in milliseconds in the above plot from the center (Case 1), top-left (Case 2), and left (Case 3) positions. The latency of the system is roughly 165 ms, which is comparable to typical human reaction times.}
  \label{fig:latencies}
\end{figure}

To quantify this delay, we developed a calibration tool that measures the end-to-end response time of the entire system. The tool renders a visual marker---AprilTags with IDs from 0 to 17---on the screen to command a specific action. As the tool picks a visual marker to render, it also starts a high-resolution timer. The robot, perceiving this marker through its camera, actuates the joystick to perform the commanded action. The tool detects the physical actuation by monitoring the electrical signals on the controller's wires. As soon as it has read the commanded action, it stops the timer.

This process measures the total response time of the system, encompassing the time taken for display rendering, camera capture, image processing, and mechanical actuation. We report the response time for various action pairs in \autoref{fig:latencies}. Each reported measurement is the average of 30 trials. The response time is roughly 165 ms, which is comparable to typical human reaction times. This response time does not include the time it would take to pick an action using a deep neural network.
\section{Experimental Setup: Network Architecture and the Learning Algorithm}
We used the Physical Atari platform with a deep reinforcement learning algorithm implemented in the \textit{Reactive Reinforcement Learning} formulation proposed by Travnik et al. (2018). In this framework, the agent sends the action to the environment as soon as it is done picking the action. The learning update happens after the action is sent and before the next observation is perceived.

The agent uses a fully convolutional network that progressively downsamples its visual input. Each camera frame is resized to $128 \times 128 \times 3$ and blended into an exponential moving average before the network receives it. The agent then normalizes each smoothed observation by subtracting its mean pixel value and dividing by its standard deviation.

The smoothed observation passes through a sequence of $3\times3$ convolutional layers (LeCun et al., 1989). After each convolution, except the final one, the feature maps are downsampled with a $3\times3$ max-pooling operation with stride 2 and then passed through a ReLU nonlinearity (Nair \& Hinton, 2010). This process continues until the spatial resolution is reduced to $2\times2$, and a linear layer maps the final feature map to a scalar value. The first convolutional layer has 24 filters, and we double the number of filters after each convolution.

\begin{figure}[t]
  \centering
  \includegraphics[width=\textwidth]{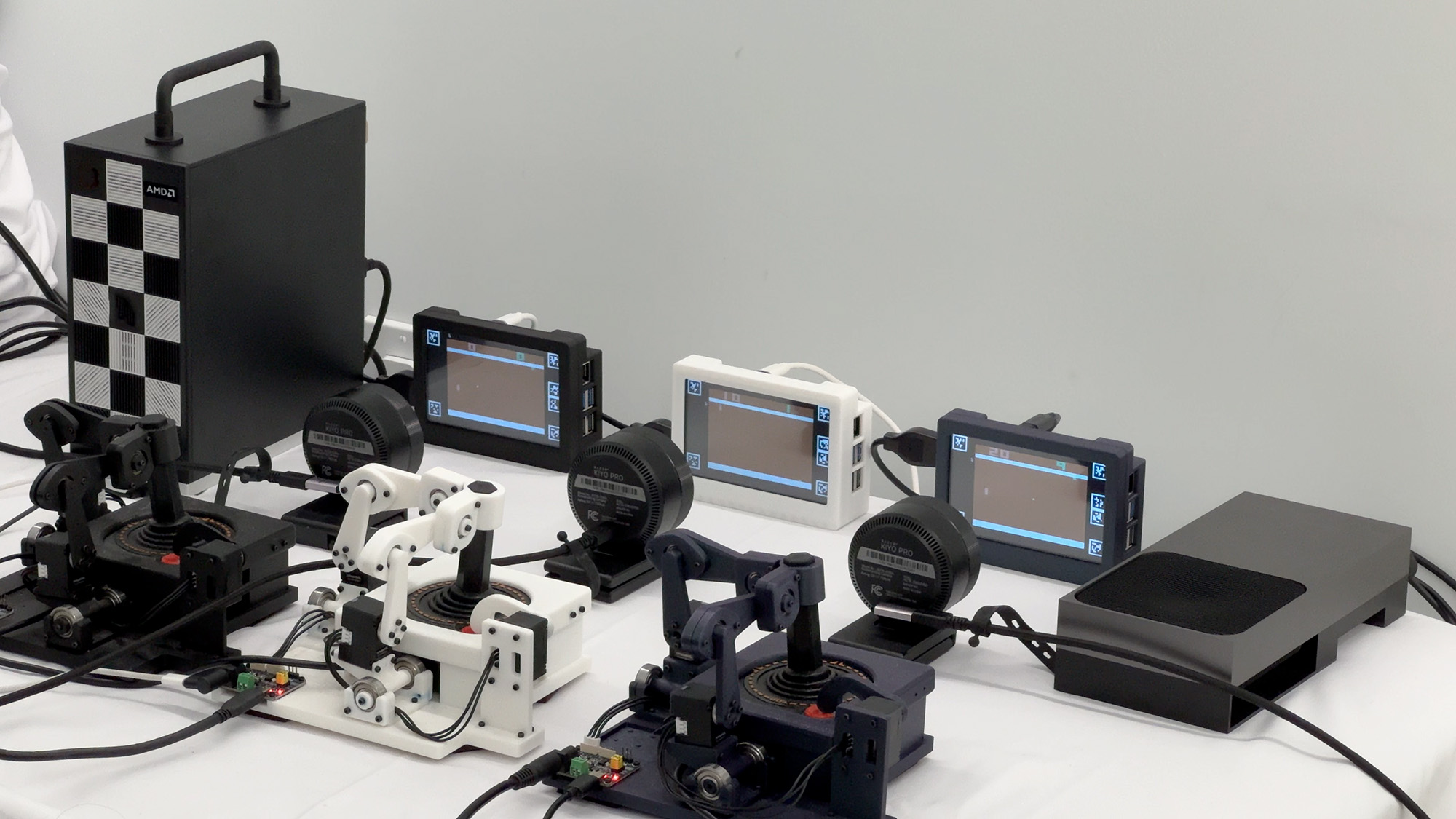}
  \caption{A photo of three Physical Atari setups.}
  \label{fig:three_robots}
\end{figure}

A distinguishing feature of this architecture is the late fusion of the agent's action history. The agent encodes its recent actions as a vector and uses an exponential moving average of that vector to summarize recent control decisions. This gives the network a compact description of what the agent has been doing over the recent past.

For Atari, we represent each action with a factored binary code. This code separates the action into vertical movement, horizontal movement, and fire, so related actions can share structure in their representation. The first three bits encode vertical movement, the next three bits encode horizontal movement, and the final two bits encode whether the fire button is pressed. For example, \texttt{UP} is encoded as \texttt{[1,0,0, 0,1,0, 1,0]} because it selects up, no horizontal movement, and no fire. \texttt{UPRIGHT} is encoded as \texttt{[1,0,0, 0,0,1, 1,0]} because it keeps the same up and no-fire components but changes the horizontal component from neutral to right.

We combine this action history with the visual stream after the first convolutional stage and immediately before the second convolution. Each entry of the action vector is copied to every spatial location in the current convolutional representation, which produces one constant feature map per action component. We then concatenate these feature maps with the visual feature maps along the channel dimension.

This design allows each convolutional filter to condition on the same recent-action summary at every spatial location. In this respect, the architecture differs from DQN (Mnih et al., 2015) and its variants, where the network outputs a value for each action instead of receiving the action history as part of its input. As a result, our value estimator is explicitly conditioned on both a temporally smoothed visual observation and the recent sequence of actions.

The targets are constructed using $n$-step returns (Sutton, 1988; Sutton \& Barto, 2018). The agent uses an $\epsilon$-greedy behavior policy with a fixed value of $\epsilon$.

The agent also maintains a separate target network. It uses this target network to evaluate actions when it selects behavior and to compute the bootstrap value at the end of each $n$-step return. We update the target network by copying the online network into it at fixed intervals.
\begin{figure}[t]
  \centering
  \includegraphics[width=\textwidth]{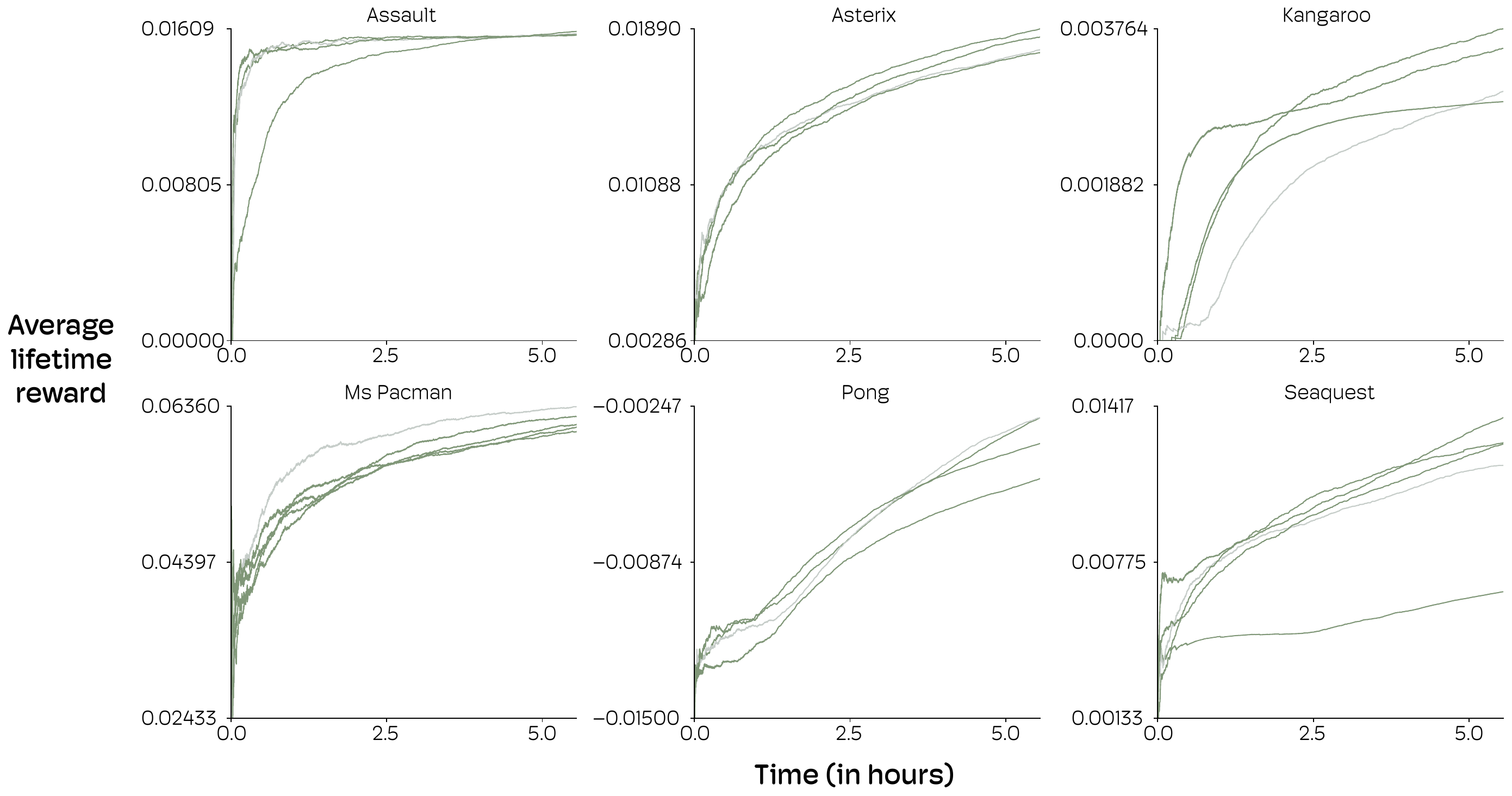}
  \caption{Average reward over time during real-time reinforcement learning on six Atari games. At every point in time, the y-axis plots the cumulative reward divided by total steps. The agent runs at 30 frames per second. On each game, we repeated the experiment 4 to 5 times with the same hyperparameters. Collectively, these experiments took around 145 hours and required no human intervention.}
  \label{fig:avg_reward}
\end{figure}

The agent learns from samples of prior experience stored in a replay buffer. Instead of sampling uniformly from the full buffer, it reserves a fraction of the batch for the most recent online transitions and samples the remaining elements of the batch randomly from the buffer. The effectiveness of this modification has been shown by Zhang \& Sutton (2017).

The agent updates its weight parameters to minimize the mean squared error between the network's predictions and the multi-step return targets. It uses AdamW (Kingma \& Ba, 2015; Loshchilov \& Hutter, 2019) to update the parameters in the convolutional layers, and uses SGD with momentum to update the parameters in the final linear layer. We provide more details of the experiment setup and the hyperparameters used in the experiments in \autoref{app:hyperparameters}. The code is available \href{https://github.com/Keen-Technologies/physical-atari-rlc}{here}.

The performance of the agent is measured as the average reward over all past experience. That is, if $r_i$ is the scalar reward at the $i$th time step, then the average reward up to time step $t$ is
$$ \frac{1}{t}\sum_{i=1}^t r_i.$$
We did not use episodic returns as an evaluation metric because average reward is a better metric for continuing reinforcement learning (Naik \& Sutton, 2019).
\section{Experiment Results}
We conducted two experiments to characterize the performance of the Physical Atari platform. The first experiment tests whether existing reinforcement learning algorithms can consistently learn on the platform without mechanical failures or human intervention. The second experiment tests how well policies learned on one Robotroller transfer to a different Robotroller.
\subsection{Physical Atari Can Be Used for Weeks of Non-Stop Learning Without Mechanical Failures}

\begin{figure}[t]
  \centering
  \includegraphics[width=\textwidth]{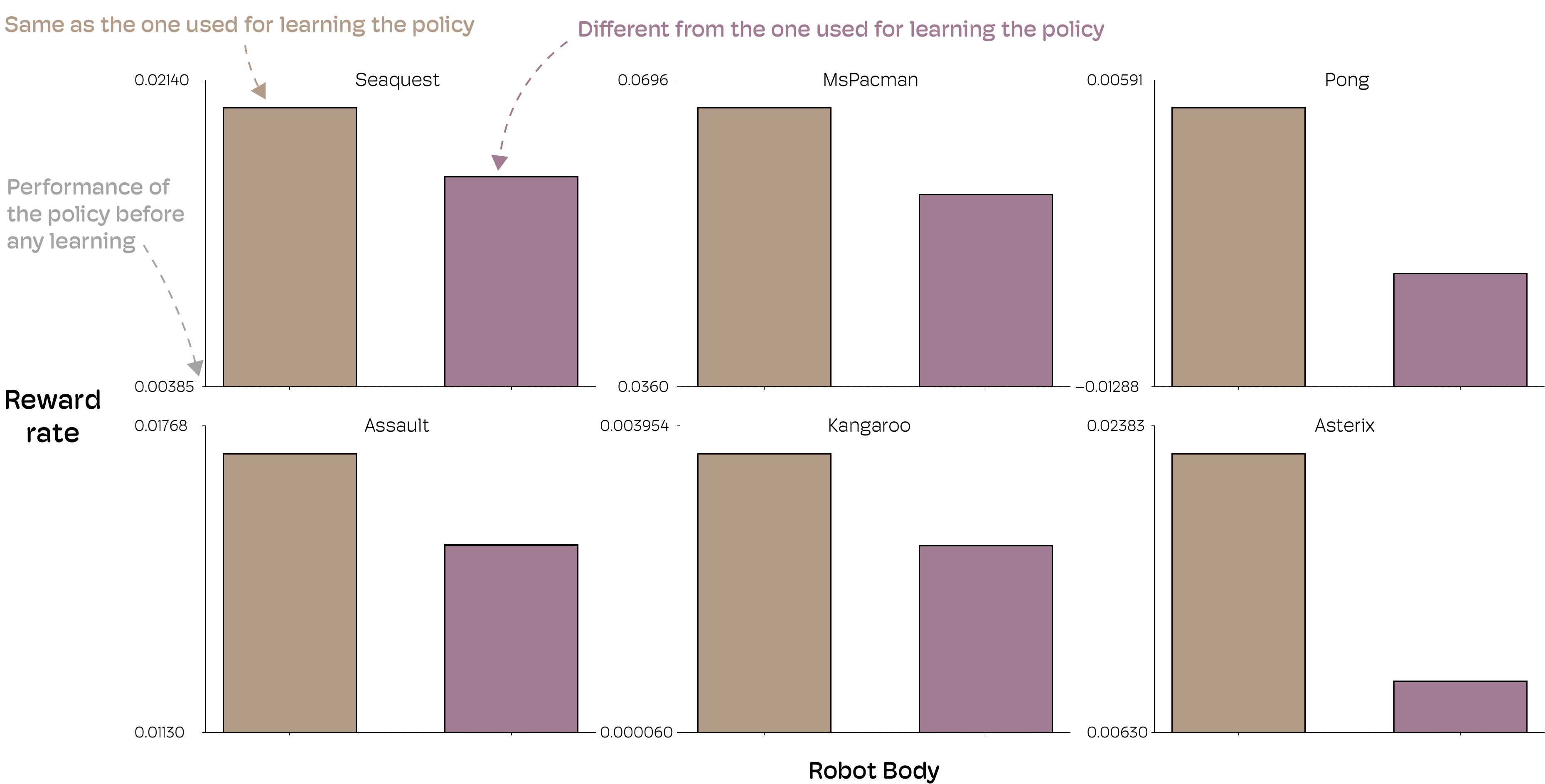}
  \caption{We evaluated policies that learned with 6 hours of experience twice, once on the robot body used for learning and once on a different but identically built body. On all six games, the performance was measurably better when tested on the body used for learning. This shows that even very small distribution shifts can negatively impact performance and that learning directly on the individual robot body can be advantageous.}
  \label{fig:real2real}
\end{figure}

We ran our agent on the Physical Atari platform to learn to play six games---Pong, Seaquest, MsPacman, Assault, Asterix, and Kangaroo---for five and a half hours. For each game, we repeated the experiment at least 4 times and report the results in \autoref{fig:avg_reward}.

The agent learned on all games, and its performance was consistent across multiple runs. These experiments took nearly 145 hours and required no interventions. A video of the system learning is \href{https://keenagi.com/research/physical-atari/}{here}.
\subsection{Small Distribution Shifts Can Have Large Negative Impacts on Performance in Physical Atari}
An appeal of the Physical Atari platform is that it can be used to study questions that often arise in robotics. One such question is how small differences in different robot bodies impact performance.
\begin{figure}[t]
  \centering
  \includegraphics[width=\textwidth]{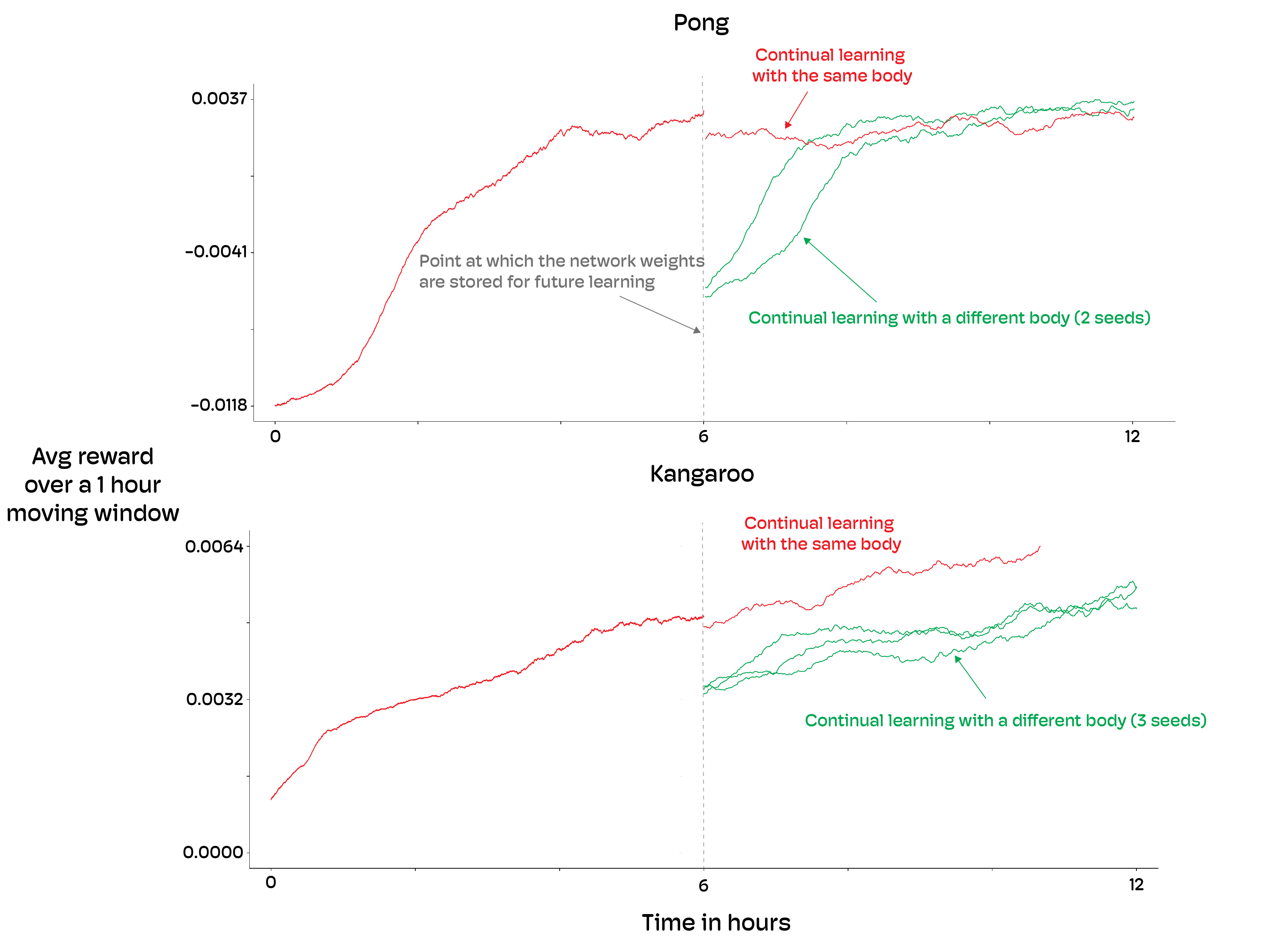}
  \caption{In another experiment, we switched the body of the robot after 6 hours of learning. The switch creates a distribution shift if the two bodies are not exactly identical. We then let the agent continue to learn with the new body. Switching the bodies degraded the performance of the policies and letting the agent learn continually improved their performance.}
  \label{fig:cl}
\end{figure}
We used our setup to answer this question. We let agents learn on six games for six hours and stored the policies. We then evaluated these policies twice, once using the Robotroller and controller that were used for learning the policies, and once using a different Robotroller and controller. We show the results in \autoref{fig:real2real}.

Across all evaluated games, the policies consistently performed worse when tested on the Robotroller that was not used for learning. The difference was largest in games that required timing actions precisely. For example, we noticed that in Pong the policy deployed on the different body moved the paddle in the right direction toward the ball but it consistently missed the ball by a small margin. Even a tiny mismatch in timing can be hugely detrimental to performance in Pong.

The degradation of performance of policies when deployed on a different body shows that even minimal distribution shifts between learning and deployment---arising from slight, inevitable physical differences between robots---can significantly impair the agent's capabilities.

We did one more experiment on two games. After deploying the policy on a new body, we let our learning algorithm continue to adapt. We show the results in \autoref{fig:cl}. Learning continually on the new body improved the policy's performance.
\section{Conclusions and Future Work}
The \textit{Physical Atari} platform is the first robotic platform, to our knowledge, that is both affordable and durable. It can be used to run experiments for weeks without requiring human intervention.

One exciting use of this platform would be to compare the three popular paradigms of RL in robotics---learning in simulation, learning from human data, and learning directly on the robots---in a systematic way. Our experiments show that even very small discrepancies between learning and deployment, introduced by using two different bodies, can lead to measurable differences in performance. Learning in simulation can introduce even larger discrepancies between learning and deployment. It would be interesting to see how well policies trained for a long time in simulation---for hundreds of millions of frames---perform when tested on the physical system. Similarly, it would be interesting to see how policies learned using a few hours of human data compare to policies learned purely from experience.

The effectiveness of the high-current reflex is another interesting result that provides a path towards more robotic platforms that are suitable for real-time reinforcement learning. Augmenting existing robotics platforms with reflexes so that a reinforcement learning agent cannot damage them could make them more suitable for real-time reinforcement learning.
\section*{Acknowledgments}
We are grateful to Sorina Lupu and Lucas Nestler for their feedback and discussions. We are also thankful to the RLC reviewers and the meta-reviewer for helpful feedback.
\appendix
\section*{References}
\hanging{Abbeel, P., Coates, A., Quigley, M., \& Ng, A. Y. (2006). An application of reinforcement learning to aerobatic helicopter flight. \textit{Advances in Neural Information Processing Systems}, 19.}

\hanging{OpenAI, Akkaya, I., Andrychowicz, M., Chociej, M., Chiek, M., Boby, A., Baker, B., ... \& Zaremba, W. (2019). Solving Rubik's cube with a robot hand. \textit{arXiv Preprint arXiv:1910.07113}.}

\hanging{Benbrahim, H., Doleac, J., Franklin, J., \& Selfridge, O. (1992, June). Real-time learning: A ball on a beam. In \textit{International Joint Conference on Neural Networks}.}

\hanging{Brockman, G., Cheung, V., Pettersson, L., Schneider, J., Schulman, J., Tang, J., \& Zaremba, W. (2016). OpenAI gym. \textit{arXiv Preprint arXiv:1606.01540}.}

\hanging{Bellemare, M. G., Naddaf, Y., Veness, J., \& Bowling, M. (2013). The Arcade Learning Environment: An evaluation platform for general agents. \textit{Journal of Artificial Intelligence Research}.}

\hanging{Beattie, C., Leibo, J. Z., Teplyashin, D., Ward, T., Wainwright, M., Küttler, H., ... \& Petersen, S. (2016). DeepMind Lab. \textit{arXiv Preprint arXiv:1612.03801}.}

\hanging{Haarnoja, T., Ha, S., Zhou, A., Tan, J., Tucker, G., \& Levine, S. (2019, June). Learning to walk via deep reinforcement learning. In \textit{Proceedings of Robotics: Science and Systems}.}

\hanging{Hessel, M., Modayil, J., van Hasselt, H., Schaul, T., Ostrovski, G., Dabney, W., Horgan, D., Piot, B., Azar, M., \& Silver, D. (2018). Rainbow: Combining improvements in deep reinforcement learning. \textit{Proceedings of the AAAI Conference on Artificial Intelligence}.}

\hanging{Kingma, D. P., \& Ba, J. (2015). Adam: A method for stochastic optimization. \textit{International Conference on Learning Representations}.}

\hanging{LeCun, Y., Boser, B., Denker, J. S., Henderson, D., Howard, R. E., Hubbard, W., \& Jackel, L. D. (1989). Backpropagation applied to handwritten zip code recognition. \textit{Neural Computation}.}

\hanging{Loshchilov, I., \& Hutter, F. (2019). Decoupled weight decay regularization. \textit{International Conference on Learning Representations}.}

\hanging{Mahmood, A. R., Korenkevych, D., Komer, B. J., \& Bergstra, J. (2018, October). Setting up a reinforcement learning task with a real-world robot. In \textit{2018 IEEE/RSJ International Conference on Intelligent Robots and Systems (IROS). IEEE.}}

\hanging{Mandlekar, A., Zhu, Y., Garg, A., Booher, J., Spero, M., Tung, A., ... \& Fei-Fei, L. (2018, October). RoboTurk: A crowdsourcing platform for robotic skill learning through imitation. In \textit{Conference on Robot Learning}. PMLR.}

\hanging{Mnih, V., Kavukcuoglu, K., Silver, D., Rusu, A. A., Veness, J., Bellemare, M. G., ... \& Hassabis, D. (2015). Human-level control through deep reinforcement learning. \textit{Nature}.}

\hanging{Naik, A., Shariff, R., Yasui, N., Yao, H., \& Sutton, R. S. (2019). Discounted reinforcement learning is not an optimization problem. \textit{arXiv Preprint arXiv:1910.02140}.}

\hanging{Nair, V., \& Hinton, G. E. (2010). Rectified linear units improve restricted Boltzmann machines. In \textit{Proceedings of the 27th International Conference on Machine Learning}.}

\hanging{Olson, E. (2011, May). AprilTag: A robust and flexible visual fiducial system. In \textit{2011 IEEE International Conference on Robotics and Automation}.}

\hanging{Ramstedt, S., \& Pal, C. (2019). Real-time reinforcement learning. \textit{Advances in Neural Information Processing Systems}, 32.}

\hanging{Schrittwieser, J., Antonoglou, I., Hubert, T., Simonyan, K., Sifre, L., Schmitt, S., ... \& Silver, D. (2020). Mastering Atari, Go, Chess and Shogi by planning with a learned model. \textit{Nature}.}

\hanging{Schwarzer, M., Obando-Ceron, J., Courville, A., Bellemare, M. G., Agarwal, R., \& Castro, P. S. (2023). Bigger, better, faster: Human-level Atari with human-level efficiency. In \textit{Proceedings of the 40th International Conference on Machine Learning}.}

\hanging{Smith, L., Kostrikov, I., \& Levine, S. (2023, July). Demonstrating a walk in the park: Learning to walk in 20 minutes with model-free reinforcement learning. In \textit{Proceedings of Robotics: Science and Systems}.}

\hanging{Sutton, R. S. (1988). Learning to predict by the methods of temporal differences. \textit{Machine Learning}.}

\hanging{Sutton, R. S., \& Barto, A. G. (2018). \textit{Reinforcement Learning: An Introduction}. MIT Press.}

\hanging{Travnik, J. B., Mathewson, K. W., Sutton, R. S., \& Pilarski, P. M. (2018). Reactive reinforcement learning in asynchronous environments. \textit{Frontiers in Robotics and AI}.}

\hanging{Van Hasselt, H. P., Hessel, M., \& Aslanides, J. (2019). When to use parametric models in reinforcement learning? \textit{Advances in Neural Information Processing Systems}, 32.}

\hanging{Wu, P., Escontrela, A., Hafner, D., Abbeel, P., \& Goldberg, K. (2023, March). Daydreamer: World models for physical robot learning. In \textit{Conference on Robot Learning}. PMLR.}

\hanging{Zhao, T. Z., Kumar, V., Levine, S., \& Finn, C. (2023, July). Learning fine-grained bimanual manipulation with low-cost hardware. In \textit{Proceedings of Robotics: Science and Systems}.}

\hanging{Zhang, S., \& Sutton, R. S. (2017). A deeper look at experience replay. \textit{arXiv Preprint arXiv:1712.01275}.}

\beginSupplementaryMaterials
\section{Atari Devbox Technical Details}
\label{app:devbox}

The Atari Devbox serves as the interface between the physical world and the emulated Atari environment. It is designed to be low-latency, reliable, and reproducible using off-the-shelf components. Note that the Atari Devbox is not used for running the learning agent code.

\subsection{Software Architecture}
The core software is a C++ application running on a Raspberry Pi 5 with Debian 12 (Bookworm). It integrates several key libraries to ensure high performance. The Arcade Learning Environment (ALE) provides the underlying Atari 2600 emulation, while SDL2 handles high-performance 2D rendering of the game frame and AprilTags. To ensure minimal input lag, the application uses `libgpiod`, a fast, modern C++ interface for reading GPIO pins. Additionally, Dear ImGui is used for debugging overlays. The main loop runs at 60 Hz, synchronized with the game emulation. In each cycle, the system reads the GPIO pins, steps the ALE environment, and renders the new frame and auxiliary tags. The GPIO pins are read just before sending the action to the ALE emulator. This ensures that the system uses the latest action.

\subsection{GPIO Interface and Action Mapping}
The Atari CX40+ joystick is a device with five wires connected to a female DB9 connector. We connect these wires to the Raspberry Pi's GPIO pins (see Table \ref{tab:gpio_mapping}). The internal pull-up resistors are enabled, so a button press pulls the corresponding pin to ground (logic 0). The raw GPIO states are mapped to the 18 discrete actions defined by the ALE. For example, if both ``Up'' (Pin 17) and ``Right'' (Pin 24) are active, the system sends the \texttt{PLAYER\_A\_UPRIGHT} action to the emulator. This allows the physical joystick to access the full action space of the Atari 2600.

\begin{table}[h]
\centering
\caption{GPIO Pin Mapping for CX40+ Controller}
\label{tab:gpio_mapping}
\begin{tabular}{ll}
\hline
\textbf{Function} & \textbf{GPIO Pin (BCM)} \\ \hline
Up                & 17                      \\
Down              & 27                      \\
Left              & 22                      \\
Right             & 24                      \\
Fire              & 23                      \\ \hline
\end{tabular}
\end{table}

\subsection{Visual Interface and Reward Communication}
To enable the agent to perceive the environment and reward solely through vision, the Atari Devbox renders a composite image containing the game screen and several AprilTags. Four static AprilTags (IDs 0--3) are rendered at the corners of the screen to allow the agent to compute a homography and extract the game frame, regardless of the camera's precise position or angle. Since the agent observes the world through a camera, the reward signal must be encoded visually. We use two dynamic AprilTags positioned on the right side of the screen to transmit reward information. A Value Tag at the top indicates the value of the reward: ID 10 for zero, ID 11 for 1, and ID 12 for -1 rewards. A Change Indicator Tag at the bottom distinguishes between consecutive rewards by cycling through IDs 15, 16, and 17 every time a non-zero reward is received. If the reward is zero, this tag remains static. This state-change mechanism ensures that the agent can accurately count the number of reward events even if they occur in rapid succession.

\section{Input/Output C++ Library}
\label{app:cpp_library}

To enable high-performance control of the Robotroller, we developed a C++ library with Python bindings. This library provides a unified interface for camera capture, robot control, and environment interaction.

\subsection{Core Components}
The library consists of three main modules working in tandem. One module handles video capture from the webcam in a separate thread to ensure the most recent frame is always available, minimizing latency, while supporting configuration of parameters like focus and exposure. A second module interfaces with the Dynamixel servos, mapping the 18 discrete Atari actions to specific servo positions using a PID controller built into Dynamixels and safety features like current monitoring. Finally, a third system utilizes the AprilTag library to detect tags in the camera feed, serving two critical functions: image rectification to crop the game screen, and reward extraction to decode the visual reward signal.

\subsection{Environment Interface}
The library integrates these components into a cohesive environment, exposing a \texttt{send\_action} method and a \texttt{perceive} method. The \texttt{send\_action} method sends the specified action to the Robotroller to trigger servo movement. The \texttt{perceive} method captures the latest frame from the camera and detects tags to rectify the image and extract the reward. This design abstracts away the complexity of the physical hardware, allowing reinforcement learning agents to interact with the Physical Atari platform using a simple API.

\subsection{High-current Reflex}
To ensure long-term reliability and prevent mechanical failure, the library implements a real-time safety reflex within the servo control loop. This mechanism protects against overcurrent events, which typically indicate that the robot is stalled or pushing against a hard limit. An overcurrent event can also put the Dynamixel into a state that requires a manual reboot of the servos. The following logic is executed continuously:

\begin{verbatim}
for (auto servo : list_of_servos) {
    int16_t cur = servo->getPresentCurrent();
    if (std::abs(cur) > 1200) {
        // Stop movement by setting target to current position
        servo->setPosition(servo->getPresentPosition());

        // Cycle torque to release mechanical tension
        servo->disableTorque();
        std::this_thread::sleep_for(std::chrono::milliseconds(1));
        servo->enableTorque();
    }
}
\end{verbatim}

When the current magnitude exceeds the threshold of 1200 mA, the system sets the target position to the servo's current position to halt movement and momentarily toggles the torque.

\section{Experiment Harness}
\label{app:harness}

To conduct reproducible experiments, we developed a unified Python harness running on the Framework computer. This harness orchestrates the interaction between the agent and the physical environment, manages data logging, and handles experiment configuration. It interfaces directly with the hardware using the Python bindings to the C++ library described in Appendix \ref{app:cpp_library}.

\subsection{Training Loop}
The core training loop is designed to minimize latency and ensure consistent timing. In each iteration, the harness first calls \texttt{env.perceive()} to update the agent state from the sensors (camera and AprilTags). It then retrieves the latest observation and reward. The agent's \texttt{get\_action} method is called with the current observation to select the next action, which is immediately sent to the environment via \texttt{env.send\_action(action)} to trigger actuation. Finally, the agent's \texttt{learn} method is invoked to update its policy based on the transition.

\subsection{Data Logging and Reproducibility}
The harness automatically logs comprehensive experiment data to a MongoDB database. All hyperparameters, random seeds, and environment settings are stored with the results. Training metrics, including average reward and episode scores, are tracked in real-time. Additionally, model weights and optimizer states are serialized and stored using GridFS, enabling exact resumption of experiments and post-hoc analysis. The harness code is available \href{https://github.com/Keen-Technologies/physical-atari-rlc}{here}.

\section{Hyperparameters}
\label{app:hyperparameters}

Table \ref{tab:hyperparameters} lists the hyperparameters used for the experiments. All games use the same set of hyperparameters.

\begin{table}[h]
\centering
\caption{Hyperparameters used in the experiments.}
\label{tab:hyperparameters}
\begin{tabular}{ll}
\hline
\textbf{Hyperparameter} & \textbf{Value} \\ \hline
Observation Size & $128 \times 128$ \\
Observation Channels & 3 (RGB) \\
Convolutional kernels in the first layer & 24 \\
Policy Skip & 2 \\
Train Skip & 4 \\
Observation EMA ($2^{\text{log}_2}$) & $2^{-3}$ \\
Momentum & 0.9 \\
Learning Rate ($2^{\text{log}_2}$) & $2^{-14}$ \\
Linear Learning Rate ($2^{\text{log}_2}$) & $2^{-17}$ \\
Exploration ($2^{\text{log}_2}$) & $2^{-6}$ \\
Target Model Update Period ($2^{\text{log}_2}$) & $2^{5}$ \\
Train Batch Size & 16 \\
Train Repetitions & 1 \\
Multi-step Returns ($n$) & 12 \\
Discount Factor ($\gamma$) & 0.99 \\
Online Samples & 4 \\
Action Encoding & 1 \\
Action EMA Rate & 0.875 \\
Action Steps & 4 \\
Action Layers & 2 \\
Max Frames Without Reward & 18000 \\
Total Steps & 600000 \\
Evaluation Steps & 50000 \\
FPS & 30 \\ \hline
\end{tabular}
\end{table}

The hyperparameters dictate how the agent perceives the environment, processes actions, and updates its policy. The \textbf{Observation Size} of $128 \times 128$ defines the resolution to which the input frames are downsampled. To smooth visual inputs and reduce noise, an exponential moving average (EMA) is applied to the observations with a rate determined by \textbf{Observation EMA} ($2^{-3}$). Similarly, the agent maintains an EMA of its action history using the \textbf{Action EMA Rate} (0.875). This action history is injected into the convolutional network at specific depths defined by \textbf{Action Layers} (bitmask 2), allowing the agent to condition its value estimates on recent decisions. The \textbf{Action Encoding} parameter (1) specifies that actions are represented using a factored binary encoding rather than a one-hot vector.

The learning process is governed by $n$-step returns, where $n$ is set to 12, meaning the target value aggregates rewards over the next 12 frames before bootstrapping. Future rewards are discounted by a \textbf{Discount Factor} of 0.99 per frame. The agent updates its weights every \textbf{Train Skip} (4) frames using a \textbf{Train Batch Size} of 16. Crucially, to ensure the agent learns from its most immediate experiences, \textbf{Online Samples} (4) of the batch are forced to be the most recent transitions, while the rest are sampled uniformly from the replay buffer.

The convolutional layers are updated using AdamW with a \textbf{Learning Rate} of $2^{-14}$, while the final linear layer is updated using SGD with \textbf{Momentum} (0.9) and a smaller \textbf{Linear Learning Rate} of $2^{-17}$. A target network is hard-updated every \textbf{Target Model Update Period} ($2^5 = 32$) frames. Finally, exploration is managed via an $\epsilon$-greedy strategy with a fixed $\epsilon$ of $2^{-6}$ (\textbf{Exploration}), and new actions are selected every \textbf{Policy Skip} (2) frames.

\section{Robotroller and Camera Hyperparameters}
\label{app:robot_camera_hyperparameters}

This section details the specific configuration used for the Robotroller and the camera in our experiments.

\subsection{Camera Configuration}
The camera is configured with manual focus and exposure to ensure consistent visual observations across frames, preventing auto-adjustment artifacts. The camera captures frames at a resolution of ($1280 \times 720$). Table \ref{tab:camera_config} lists the specific parameters.

\begin{table}[h]
\centering
\caption{Camera Configuration Parameters.}
\label{tab:camera_config}
\begin{tabular}{ll}
\hline
\textbf{Parameter} & \textbf{Value} \\ \hline
Camera Resolution & $1280 \times 720$ \\
Camera FPS & 30 \\
Camera Focus & 370 (Manual) \\
Camera Zoom & 100 \\
Camera Exposure & 20 (Manual) \\
Camera Brightness & 128 \\
Camera Contrast & 128 \\ \hline
\end{tabular}
\end{table}

\subsection{Robot Configuration}
The Robotroller's movement is governed by a set of parameters that control communication and motor behavior. The \textbf{Baud Rate} of 1,000,000 sets the speed of serial communication between the computer and the Dynamixel servos.

The servos use a Proportional-Integral-Derivative (PID) controller to precisely reach and hold target positions. The \textbf{P gain parameter} (1500) determines how aggressively the motor reacts to position error; a higher value increases speed but may cause overshoot. The \textbf{I gain parameter} (6000) corrects for steady-state error, ensuring the joystick is held firmly against resistance. The \textbf{D gain parameter} (1500) dampens the movement by reacting to the rate of change, reducing oscillation and ensuring smooth motion.

The servo positions are calibrated specifically for the robot to map the discrete Atari actions to precise physical encoder values (0-4095). \textbf{Noop Position (Default)} (2048) represents the neutral position. \textbf{Right/Left Position} and \textbf{Up/Down Position} define the target values for pushing the joystick in the respective directions. \textbf{Button Default Position} (2000) and \textbf{Button Pressed Position} (1932) define the released and actuated states of the fire button, respectively. Table \ref{tab:robot_config} lists these parameters.

\begin{table}[h]
\centering
\caption{Robotroller Configuration Parameters.}
\label{tab:robot_config}
\begin{tabular}{ll}
\hline
\textbf{Parameter} & \textbf{Value} \\ \hline
Baud Rate & 1,000,000 \\
P gain parameter & 1500 \\
I gain parameter & 6000 \\
D gain parameter & 1500 \\
Noop Position (Default) & 2048 \\
Right Position & 2130 \\
Left Position & 1925 \\
Up Position & 2180 \\
Down Position & 1960 \\
Button Default Position & 2000 \\
Button Pressed Position & 1932 \\ \hline
\end{tabular}
\end{table}

\end{document}